\documentclass[10pt,conference]{IEEEtran}
\usepackage{amsmath,graphicx,comment}
\usepackage{relsize}

\usepackage[hyphens]{url}
\usepackage[mathletters]{ucs}
\usepackage[utf8x]{inputenc}
\usepackage{xcolor}

\title{Whole page recognition of historical handwriting}
\usepackage[keeplastbox]{flushend}

   \author{
        \IEEEauthorblockN{Hans J.G.A. Dolfing}
        \IEEEauthorblockA{ \textit{Independent}\\
       	 	Washington, DC (USA)\\
        	leovinus@protonmail.com}
    } 

\begin{document}
%
\maketitle
\baselineskip 11pt
\begin{abstract}
Historical handwritten documents guard an important part of human knowledge only within reach of a few scholars and experts. Recent developments in machine learning and handwriting research have the potential of rendering this information accessible and searchable to a larger audience. 
To this end, we investigate an end-to-end inference approach without text localization which takes a handwritten page and transcribes its full text. No explicit character, word or line segmentation is involved which is why we call this approach ``segmentation free''.
We explore its robustness and accuracy compared to a line-by-line segmented approach based on the IAM, RODRIGO and ScribbleLens corpora, in three languages with handwriting styles spanning 400 years. We concentrate on model types and sizes which can be deployed on a hand-held or embedded device. We conclude that a whole page inference approach without text localization and segmentation is competitive.
\end{abstract}


\section{Introduction}
\label{sec:intro}

In the context of handwriting recognition applied to historical handwritten manuscripts, we investigate an approach to improve and simplify the recognition of whole pages of handwritten text. This work fits in the wider field of competitions on historical documents, document layout and processing, text-in-the-wild challenges and document segmentation. 

We concentrate on an end-to-end approach to avoid unrecoverable, early errors due to segmentation or localization. Due to the nature of the old manuscripts, we focus on off-line handwriting recognition based on static archive images~\cite{Nakagawa04, Steinherz99, Vincarelli02}. 

When computational resources were limited, divide-and-conquer techniques were important to split the system level processing into manageable parts. Segmentation is such a divide-and-conquer approach which has been used in the field for decades from the segmentation of characters into strokes and radicals~\cite{Nakagawa04}, words into characters~\cite{Bozinovic89}, sentences into words, and pages into sentences and other content~\cite{Mehri19}. 

With the improvement of CPUs and algorithms, segmentation free approaches have started to take over, e.g., recognition of cursive handwritten words via hidden Markov models (HMM)~\cite{Dolfing98} and neural networks~\cite{Senior94, Manke94}, full handwritten sentences~\cite{Starner94} and even paragraphs~\cite{BlucheLM16}. In the field of speech recognition, development is similar from Bayesian continuous word and sentence recognition~\cite{Ney84} to recent jointly-trained, end-to-end speech recognition approaches~\cite{Wang18}.

\begin{figure}[htbp]
	\centering
	\includegraphics[scale=0.2]{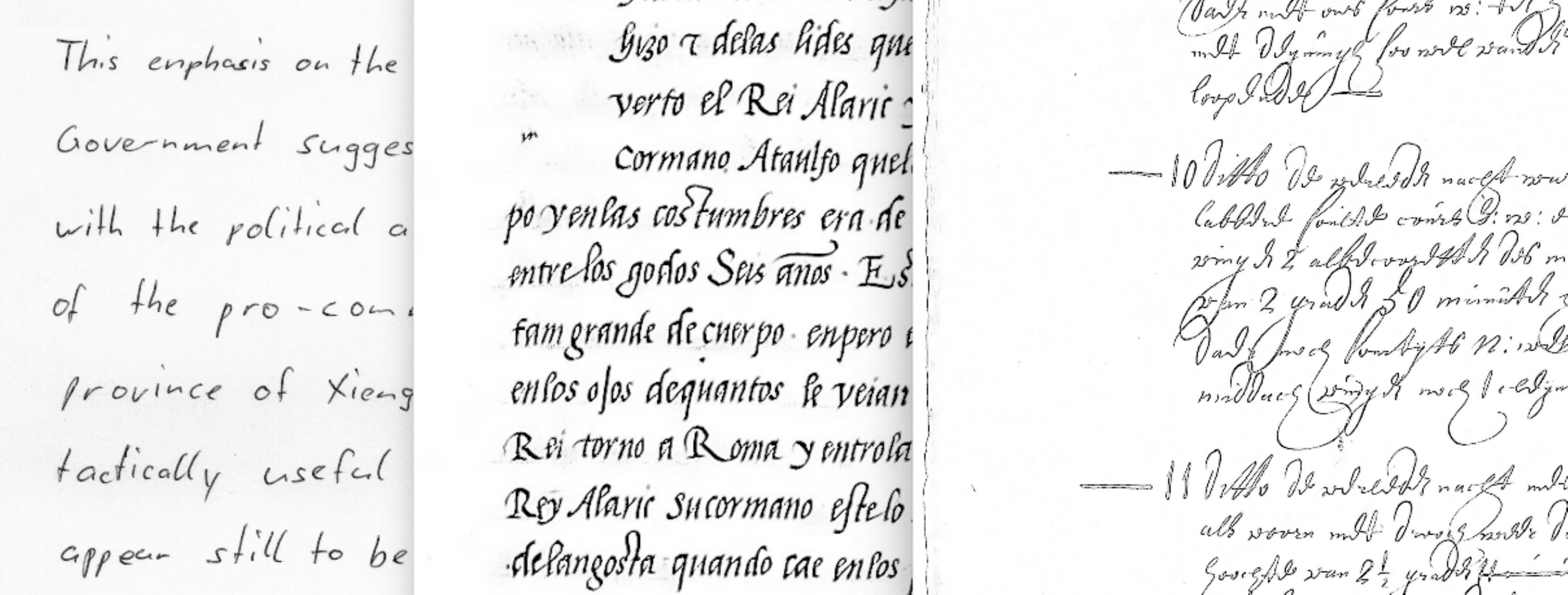} 
	\caption{Example pages IAM, RODRIGO, ScribbleLens}
	\label{fig:examplepages}
\end{figure}

In this paper, we investigate a segmentation-free inference approach for whole pages of handwritten historical manuscripts such as in Figure~\ref{fig:examplepages}. With the goal of a simple and effective recognition system, we prototype our approach on the IAM~\cite{Bunke99}, RODRIGO~\cite{Rodrigo10} and ScribbleLens~\cite{Dolfing20} corpora. 

In Section~\ref{sec:previous} and Section~\ref{sec:corpora}, we discuss the related work and corpora. Section~\ref{sec:setup} contains details on the experimental setup comprising models and approaches. Results are presented in Section~\ref{sec:result} followed by the conclusion in Section~\ref{sec:conclusion}.

\section{Related work}
\label{sec:previous}

Recognition architectures on image-based, off-line handwriting~\cite{Vinciarelli02} have evolved from hidden Markov models (HMM) to deep learning-based systems with building blocks of CNN, (MD)LSTM, CTC and Transformers~\cite{Voigtlaender16,Nephi18,Graves09}. Historical or medieval off-line handwritten datasets come with an additional set of challenges which range from old morphology, orthography and grammar to image artifacts such as bleed-through, water damage, and faded ink. Current work includes comprehensive algorithm comparisons on image segmentation and localizations~\cite{Mehri19}, modeling and competitions~\cite{Sanchez17} and general overviews~\cite{Ogier08}. 

Last year, we published the ScribbleLens~\cite{Openslr19,Dolfing20} corpus with historical, handwritten, Dutch manuscripts. We investigated supervised and unsupervised representation learning~\cite{Lancucki2020robust}. The work here takes a different tack on the same corpus and we investigate whether we can achieve accurate recognition without explicit segmentation of pages into lines. We compare the accuracy of line-by-line, and page-by-page recognition, and investigate the robustness towards image noise and line breaks on the pages. 

Traditionally, off-line handwriting recognition concentrated on character, word and sentence images, while whole-page recognition approaches and models are more recent. Some systems follow a divide-and-conquer approach and combine separate independent components for text detection, localization and recognition, while other systems aim to combine these components in an end-to-end trainable model, just like in speech recognition~\cite{Wang18}. For the purpose of this study, a segmentation free recognition of whole pages is defined as a one-stage, end-to-end trainable system with a page image as input and transcribed text as output, without explicit stroke, character, word or line segmenter as preprocessor. 

As a first example, \cite{Shi16} consists of a combination of CNNs and BLSTMs similar to the CNNs plus Transformers and BLSTMs in this work and~\cite{Dolfing20}. The system~\cite{Shi16} is tested on images of sheet music, street views and word images. The SEE system~\cite{Bartz18} is another end-to-end trainable system which combines localization and recognition and applies it on the words from the SVHN data.

Next, the work of~\cite{Moysset17} demonstrates a full-page recognition system on noisy or heterogeneous page images. It combines detection and localization with a CNN-style regressor and an LSTM-based recognition engine. Significantly, it couples segmentation/localization plus recognition more tightly. 

Finally, the \textit{Start, Follow, Read (SFR)} model~\cite{Wigington18} is another end-to-end trainable classifier which integrates segmentation and recognition on the recognition of paragraphs and pages roughly comparable to the page-based recognizer of~\cite{BlucheLM16}. Both page-based recognizers combine localization and recognition models. In contrast, the current work only does recognition on the whole pages of historic manuscripts without prior text localization or detection.

On the flip-side, the traditional off-line printed text recognition (OCR) has seen much recent work on text localization, detection, as part of OCR text-in-the-wild work such as~\cite{Jaderberg16}. Some of these systems share the goal of an end-to-end trainable system with this paper but also address text localization which we do not use here. 

The growing list of research corpora with historical, handwritten manuscripts includes IAM-HistDB~\cite{IAM-HistDB}, the Spanish RODRIGO corpora~\cite{Rodrigo10}, GERMANA~\cite{Germana10}, READ~\cite{READ}, and the George Washington papers~\cite{Washington41}. Furthermore, thousands of personal letters in early 17th century Dutch~\cite{Wal15}, Shakespearean letters and materials~\cite{EMMO}, plus work on historical manuscript corpora in Asian languages~\cite{CASIAAHCDB19} are being bundled as research corpora for linguistics, history and paleography.

In this work, we primarily investigate the accuracy differences between classifiers trained on either segmented lines versus whole pages of historical manuscripts.
In other words, when you feed the classifier a whole page instead of segmented lines, how accurate and robust is the end-to-end transcription? 
This page-based recognition uses no lexicon, language models, text localization, text detection or line segmentation.

Secondly, we study the effect of image noise on recognition accuracy. Noise such as margin annotations, bleed-through, page numbers and other artifacts. The research question at hand is whether the accuracy degradation due to margin annotations and other noise is benign or catastrophic.

\section{Corpora}  
\label{sec:corpora}

To benchmark the proposed system, we use three off-line handwritten corpora summarized in Table~\ref{tab:corpora}. These are ScribbleLens~\cite{Openslr19,Dolfing20} in early Modern Dutch from the 16-17th century, the 16th century Spanish RODRIGO manuscript~\cite{Rodrigo10,Granell18}, and IAM~\cite{Bunke99,Bunke02} in modern English. All segmented lines and whole pages are binarized with black background and white signal. 

\begin{table}[htbp]
\begin{center}
   \caption{Corpora properties}
   \label{tab:corpora}
\resizebox{8cm}{!} {
    \begin{tabular}{| l | c | c | c | c | c |}
    \hline
		\textbf{Name} &  \textbf{nPages} & \textbf{nLines} & \textbf{nCharacters} & \textbf{nSymbols}& \textbf{Density} \\ 
					& train/test 	&	train/test	&	train/test		&					&	lines/page, char/page, char/line   \\ \hline
	IAM								& 	747/232		& 	 6161/1861		& 	285488/ 83044	& 		78		& 		 8.2  /  382.2	/ 46.3 \\
	Rodrigo							& 	619/212		& 	15092/5030		& 	845390/279862		& 	112		& 		24.3 / 1365.7	/ 56.0 \\ 
	ScribbleLens		& 	178/21	& 		 4783/563	& 		247081/ 29076	& 		80	& 			26.9 / 1388.0	/ 51.7 \\	 \hline
	\end{tabular}
}
\vspace{-0.5cm}
\end{center}
\end{table}

Character error rates (CER) are reported based on the best path and unconstrained character output. No lexicons nor language models (LM) are used to postprocess the results as our main interest is on the page-based recognition and robustness for historical handwritten manuscripts across languages. The linguistic and morphological modeling of 16th century Dutch or Spanish is an additional challenge which is why we report CER but not word error rate (WER).

The IAM off-line handwriting database~\cite{Bunke99,Bunke02} is an established corpus in modern English for off-line handwriting recognition on words and sentences, and more recently on paragraphs and pages~\cite{BlucheLM16}. We use the corpus as a sanity test and comparison of modern handwriting pages versus historical pages with additional artifacts. The writing styles vary from handprinted to very slanted and cursive. Based on the IAM provided splits~\cite{IAMSplits99}, there are 6161, 900, 940 and 1861 lines and 747, 105, 115, and 232 pages to {\em train}, {\em validate1}, {\em validate2} and {\em test} respectively, from 500+ writers and with 79 character symbols including whitespace. The latter is also included in the error counts and training. Some papers combined the {\em test} and {\em validate2} data as a common test set with marginally better CER but we use {\em test} data only to report results.

We binarize all line and page images with the provided pixel threshold levels in the IAM XML data, and experiment on both original and de-slanted handwritten lines. For recognition of words and lines, de-slanting~\cite{Scheidl18} is reported as a useful preprocessing~\cite{Voigtlaender16}. Typically, IAM recognizers use both language models and lexicons to produce the best WER and CER for words based on line image input. 
In contrast, the current work trains only on full lines and pages from its {\em lines} and {\em forms} directories and reports CER on the unconstrained symbol output from {\em test} data only.

In earlier work, IAM words and lines have been classified with CNNs and convolutions~\cite{Such16}, LSTMs~\cite{Voigtlaender16} and dropout~\cite{Pham13} as well as HMM-based~\cite{Kozielski13} and hybrid NN-HMM~\cite{Dreuw11, Bleda11} systems. Previous work which reports a CER on the unconstrained character output based on line images without use of lexicon nor LM includes a MDLSTM-RNN-Dropout system~\cite{Pham13} with 10.8\%, the MDLSTM in~\cite{BlucheLM16} reports 6.6\% and the end-to-end Start, Follow, Read approach~\cite{Wigington18} reports 6.4\%. For reference, with contextual knowledge the CER on lines input is reduced to 3.4\%~\cite{Poznanski16, Voigtlaender16}. Finally,~\cite{BlucheLM16} reports on paragraph or whole page recognition with a data setup similar to this paper and achieves 16.1\% CER on the whole pages.

The ScribbleLens~\cite{Openslr19,Dolfing20} manuscripts contain early modern Dutch, historical, handwritten, cursive text from the discovery age roughly between 1590-1750. The corpus~\cite{Openslr19} was used for supervised CER baselines~\cite{Dolfing20} as well as weakly-supervised representation learning~\cite{chorowski2019unsupervised,Lancucki2020robust}. In this paper, we extend the corpus and work to train from segmented and annotated lines as well as full pages. Based on the 200 full transcribed pages, we have 4783 plus 563 lines to train and test from a set of writers~\cite{Openslr19}. The same material is also split into 178 and 21 full page images for training and test. The train and test splits are maintained strictly, i.e., all 563 lines of test data are from the same pages as the 21 test pages. The 800 untranscribed pages are not used here.

The corpus already contains the binarized line images and its pages were binarized with the same tool as the lines. A similar procedure is used for the RODRIGO corpus. The whitespace in the transcripts is significant but never repeated. It is included in training targets and CER. In~\cite{Dolfing20}, we reported a ScribbleLens baseline CER of 11.9\% errors with augmented data~\cite{Specaug2019,Chorowski19} training. Even as the total number of writers of ScribbleLens compared to RODRIGO is significantly higher, the main difference is that the writing style in ScribbleLens is way more intricate and cursive which is traditionally much harder to recognize compared to an almost handprinted style in RODRIGO. Unlike noisy text-in-the-wild approaches~\cite{Seytre19}, the segmented lines of ScribbleLens are relatively clean but include discoloration and similar artifacts. 

The historical, handwritten RODRIGO corpus~\cite{Rodrigo10,Granell18} is handwritten by one 16th century Spanish writer annotated with line and word segmentations plus transcripts. A CNN plus RNN system~\cite{Granell18} gave CERs of about 21\% and 3\% with a 1-gram and 10-gram character LM, respectively. We speculate that without any character LM or lexicon, the CER must have been worse than 21\%. For our purposes, we re-segmented the pages of the corpus into lines with the same tool as for ScribbleLens~\cite{Openslr19}. Based on the suggested split in train and test data~\cite{Granell18} of 75 to 25 percent, we conducted our character recognition experiments on lines and full pages. The partition of pages into train and test sets is identical in both cases. The CER result on an unconstrained character output is in Section~\ref{sec:result}.

\section{Experimental setup}
\label{sec:setup}

Our goal is to build an accurate recognition model that uses a page image as input and transcribes that page as an unconstrained character sequence output without the use of an explicit line segmenter.
We perceive the software complexity reduction from segmentation plus recognition to a recognition-only approach as significant.
Unlike text-in-the-wild OCR or recognition of very heterogeneous pages, we know that journals such as RODRIGO or ScribbleLens are fairly regular and might not need further text detection or localization. 
We study the accuracy effects of linebreaks and manuscript noise such as margin annotations and bleed through. 
Other knowledge sources such as language models and grammars can always be added later.
In summary, we investigate a segmentation-free recognition approach based on a whole page image input.

Our setup is a custom trainer based on PyTorch 1.5 and derived from~\cite{Chorowski19}. The network is structured like~\cite{Dolfing20} as a CNN stack on top of a TransformerEncoder~\cite{Vaswani17,TransformerEncoder20} with a CTC loss.  We compare the TransformerEncoder with the BLSTM from~\cite{Dolfing20} to gain insight in accuracy and generalization. The network optimizer is Adam, and compute its {\em beta2} to have a momentum larger than 0.1 based on the corpus size.

As input images, we do consider the original 2D page scans and a flattened 1D version of the page to form a very long line of text with all page content~\cite{Openslr19}. We do not consider Spatial Transformer Networks~\cite{STN} for text localization, or other spatial transforms such as folding back the left and right side of page to form a cylinder while shifting the right side down one line to form a long corkscrew sentence. 
While the segmented image lines from IAM to ScribbleLens have roughly 50 characters as targets, see Table~\ref{tab:corpora},
a whole page could have 1500 target symbols or more. 

The line-by-line  input images for all three corpora are resized to dimension $[ 64 \times W]$, i.e., a standard image height of 64 and a variable width or length $W$. 
The reduced length $W'$ after CNN and before the Softmax is determined only by the number and value of strides in the CNN stack. The TransformerEncoder and BLSTM will not change the sequence length. Therefore, the sequence length after the full CNN stack and before the Softmax is typically $ W' == W/ 8( ==2^3)$.

As there are 512 output images from the CNN, we feed a sequence of  $[512 \times W']$ to 
the TransformerEncoder or BLSTM. Image padding is generally $1$.
The typical CNN settings are in Table~\ref{cnnSetup}. 
Please note the wide kernel sizes in the horizontal direction of layers $7$ and $8$ which improves accuracy significantly. 
As image height is reduced in the CNN stack step-by-step, the rationale for the wide kernels at the end of the stack is to capture more horizontal context in order to handle and exploit co-articulation.

\begin{table}[htbp]
\vspace{-0.25cm}
\begin{center}
   \caption{CNN settings}
   \label{cnnSetup}

\resizebox{8cm}{!} {

    \begin{tabular}{| l | c | c | c | c | c | c |c | c |}
    \hline
		\textbf{Layer} & \textbf{1} & \textbf{2} & \textbf{3} & \textbf{4} & \textbf{5}& \textbf{6} & \textbf{7} & \textbf{8} \\
    \hline
		Kernelsize 	& 3, 3 & 5, 5 & 3, 3 & 3, 3 & 3, 3 & 3, 3 & 3, {\bf 5} & 2, {\bf 7} \\
		Stride 			& 2, 1 & 2, 2 & 2, 2 & 1, 2 & 2, 1 & 1, 1 & 2, 1 & 1, 1 \\
		NumOutputImg  	& 64 & 128 & 128 & 256 & 256 & 512 & 512 & 512 \\
    \hline
	\end{tabular}
}
\vspace{-0.25cm}
\end{center}
\end{table}

We use a back-end of two layers of TransformerEncoders or BLSTMs plus Dropout and BN. 
Longer sequences $W'$ generally deliver better accuracy but are more difficult to train. 
In order to effectively use the CTC loss we must ensure that the sequence length after the CNN stack is larger than the amount of character symbols to be recognised ($W' >> nTargets$).
This is especially relevant for pages that have, e.g., 1500 characters as targets, meaning that we need sequences of length $W'$, e.g., 4,000 or 6,000 vectors, before the CTC.
Training on long input image sequences with long target sequences is always a bit more challenging compared to the shorter lines.

One interesting aspect of this paper is that we use the identical CNN/back-end/CTC for both line and page recognition. 
The way the page recognition works is we provide a resized input image for the whole page of size $[64*L \times W]$.
As an example, we might choose a line density scale factor $L=24$ and resize an input image from $[4000 \times 3000]$ to $[64*24 \times 1152] == [1536 \times 1152]$.
Hence we scale from $[H \times W]$ to $[64L \times (W/H) * 64L]$. 
As both height and width of the input image are scaled by $L$, the sequence length  $W'$ after the CNN and before the back-end will increase quadratically with increasing $L$.  

Per line ``density'' in Table~\ref{tab:corpora}, we choose an oversample factor of $L$ between $1$ and $32$ to build a resized page image which the CNN stack will convert to $512$ input images or tensors of $[L \times W']$ representing 512 output images after the CNN and before the back-end. Work on much denser pages with, e.g., 100 lines/page would obviously require a different $L$.
Next, we flatten and transpose that to a sequence $[ 512 \times (L * W' ) ]$ as input vectors for the back-end.
Algorithmically, the flattening to a sequence of length $L * W'$ is straightforward and surprisingly effective.

Generally, training whole pages leads to sequences in the range of about $1000$ to $12000$ with target sequences of length $O(10^3)$. The training from scratch is feasible but slow to converge. To speed up convergence, we use a form of curriculum training~\cite{Curriculum09} to bootstrap the page models. Hence, curriculum training is an engineering shortcut and not a necessity.
First, we train the usual line recognition models based on their input images of $[64 \times W]$.
Then, we use the line model with converged CNN filters to bootstrap the page model which process full page images of $[64*L \times W]$.
Page images have an unknown number of text lines roughly between 1 and 40.

We experimentally determined reasonable oversample factors $L$ for the pages 
as well as horizontal CNN strides to find setups that fit in CUDA memory for training. 
The accuracy results are reported with uncertainty computed from corpus size as discussed in~\cite{Guyon96}.

\section{Results}
\label{sec:result}

First, we compute baseline models on the line-by-line segmented data. The line image height is fixed as 64 while the line width is flexible. We compare a TransformerEncoder and BLSTM back-end which we refer to as model I and II, respectively. As a sanity test, we evaluate on IAM with a de-slanted corpus and we get a test CER of 9.10 +/- 0.33 and 7.39 +/- 0.30 for model I and II, respectively, without any language model nor lexicon which compares well to the unconstrained CER from the literature. 

Next, we re-evaluated the augmented, supervised, line-by-line ScribbleLens results from~\cite{Dolfing20} where we achieved CER 11.9 +/- 0.68. Now we achieve significantly better results of 7.00 +/- 0.49 for model I and 6.47 +/- 0.90 for model II. In the order of relevance, the improvement consists of the combination of two wider CNN kernels, see Table~\ref{cnnSetup}, larger image input height $64$ instead of $32$, and better augmentation. On RODRIGO, we achieve a CER of 2.42 and 1.93 for model I and II, respectively, which is significantly better than what was reported in the literature and discussed above. 

Table~\ref{tab:accuracy-base} summarizes results on the segmented lines of the corpora from Table~\ref{tab:corpora} and shows that both models are competitive in CER on known and new tasks.

\begin{table}[htbp]
\begin{center}
   \caption{Test CER[\%] on segmented, handwritten lines without language model or lexicon. Model size roughly between 12 and 14 million parameters.}
   \label{tab:accuracy-base}

    \begin{tabular}{| l | c | c|}
    \hline
		\textbf{Corpus}  & \textbf{Model I} & \textbf{Model II} \\  \hline  
		IAM 			& 9.10 +/- 0.33 & 7.39 +/- 0.30 \\
		RODRIGO 		& 2.42 +/- 0.09 & 1.93 +/- 0.08 \\
		ScribbleLens	& 7.00 +/- 0.49 & 6.47 +/- 0.47 \\	\hline
	\end{tabular}
\vspace{-0.25cm}
\end{center}
\end{table}


Next, we generalize the line-by-line model to a page-by-page model. As discussed earlier, we bootstrap the page-by-page model based on the line-by-line models from Table~\ref{tab:accuracy-base}. In other words, the image input to the model will change from $[64 \times W ]$ to $[64*L \times W]$ where $L$ is the vertical scaling factor that roughly denotes an expected number of lines. With ScribbleLens as example, we train and test on the page images and sweep over $L$ from 1 to 32 in Table~\ref{tab:page-sweep}. The CUDA memory usage goes up quickly with $L$ until we run out of 11GB memory after $L=32$ for this setup. Here, we chose $L=24$ as compromise in accuracy vs. CUDA memory usage and note that if we could train with larger upsampled $L>>32$ then we expect the accuracy and CER to improve further.

\begin{table}[htbp]
\begin{center}
   \caption{Page model I, ``tasman'' subset CER[\%], sweep with $L$ to identify working point for page recognition}
   \label{tab:page-sweep}
   \resizebox{8cm}{!} {

   \begin{tabular}{| l | c | c|c|c|c|c|} \hline
{\bf L}	& {\bf 1}	&	{\bf 12}	&	{\bf 18}	&	{\bf 24}	&	{\bf 25}	&	{\bf 32}		 \\ \hline
{\bf Height} &	64	&	768	&	1152&	1536&	1600&	2048	 \\ 
{\bf err[\%]}	& $\approx$ 100.0 &	64.25 &	25.30 &	11.39 &	10.54 &	11.71	 \\ 
    \hline
	\end{tabular}
}
\vspace{-0.5cm}
\end{center}
\end{table}


Now that we have a working point at $L=24$, and format all input page images to $[1536(=24*64) \times W]$, we can build page recognizers for the corpora. The result is in Table~\ref{tab:page-result} for both model types. As an excellent start, the RODRIGO page-by-page CER is very close to its line-by-line CER in Table~\ref{tab:accuracy-base}. The ScribbleLens page-by-page result is better than the line-by-line CER from last year~\cite{Dolfing20} though it seems unable to reach the line-by-line results from Table~\ref{tab:accuracy-base} in this work. For IAM page CER, the corpus setup is identical to~\cite{BlucheLM16} and we measure a slightly better CER.

\begin{table}[htbp]
\begin{center}
   \caption{Page model, test CER[\%], without language model or lexicon.}
   \label{tab:page-result}

    \begin{tabular}{| l | c | c|}
    \hline
		\textbf{Corpus} & \textbf{Model I} & \textbf{Model II} \\
    \hline
		IAM 			&  14.47 +/- 0.42 &  12.99 +/- 0.40 \\ 
		RODRIGO  		&  2.30 +/- 0.09  & 2.25 +/- 0.09\\ 
		ScribbleLens  	&  10.62 +/- 0.60 &  8.59 +/- 0.54 \\ 
    \hline
	\end{tabular}

\vspace{-0.25cm}
\end{center}
\end{table}

\begin{figure}[htbp]
	\centering
	\includegraphics[scale=0.2]{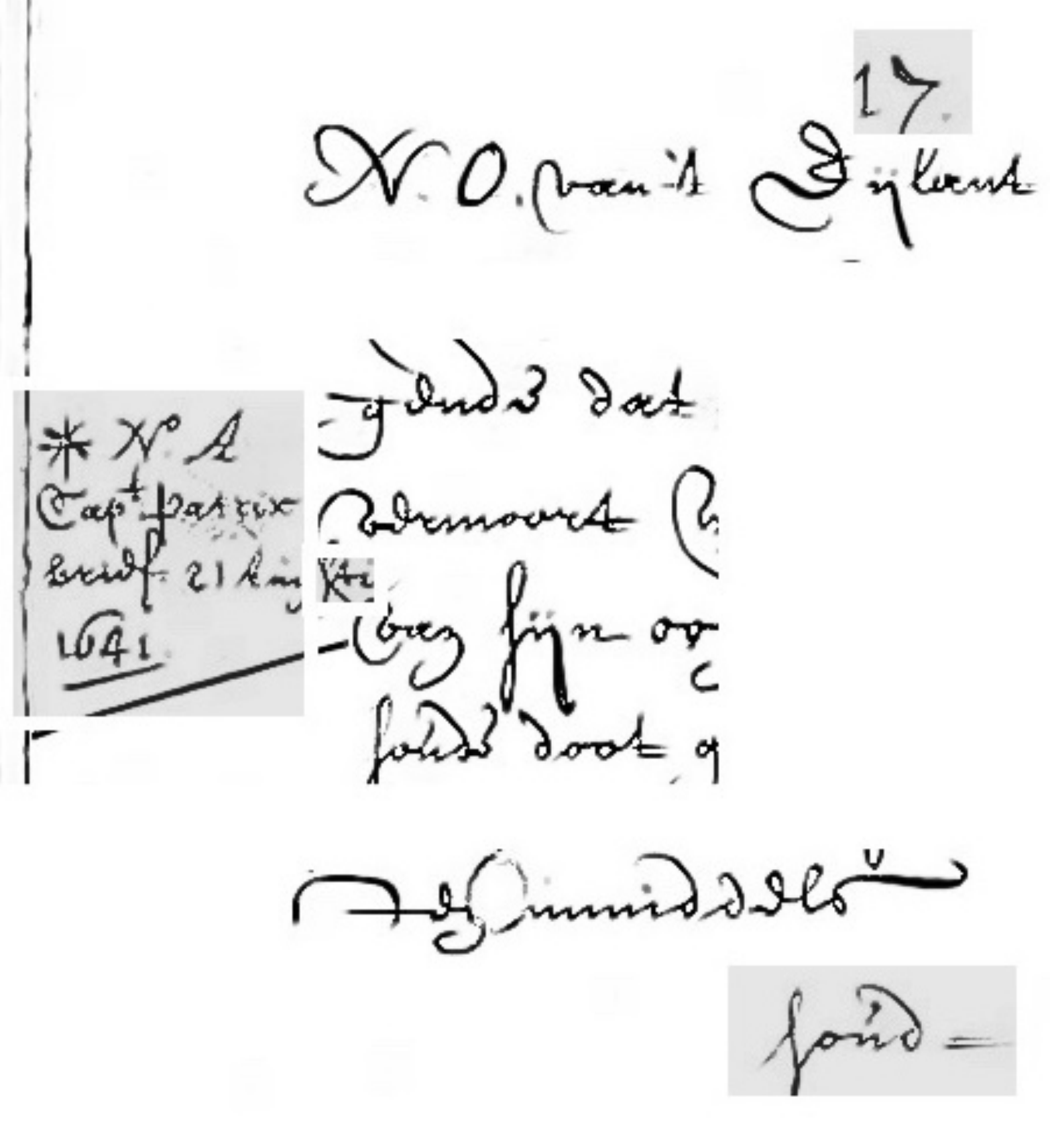} 
	\caption{Three ScribbleLens unmodelled margin annotation examples in gray, and proper handwritten text in black. From a page number top of page (right), to a section in the middle (left), and the last scribble at the bottom right.}
	\label{fig:margin}
\end{figure}

Next, we continue with model I/TransformerEncoder only and measure the robustness of the whole page recognizer with respect to noise in train and test data such as margin annotations or page numbers or similar, see Figure~\ref{fig:margin}. In other words, does the transcription accuracy change significantly when we have some unmodelled margin annotations or page numbers in the train and test data? This also relates to accuracy degradation with word-based models and recognition where there are unmodelled, out-of-vocabulary (OOV) words.

As all 1000 pages of the ScribbleLens corpus have a line-by-line segmentation, we have a way to exclude additional page numbers, additional margin annotation, either vertical or horizontal, from the manuscript pages. In contrast to the ``Original'' supervised train and test set of 200 pages from~\cite{Openslr19}, we compare to new train and test data on the newly constructed ``Clean'' 200 page set. 

The ``Clean'' dataset was constructed with a script that rebuilds each page based on the segmentation boxes, an empty page background, and copied in the same lines in the correct positions. The effect is that the new ``Clean'' supervised, corpus has minimal to no page noise whatsoever. Both ScribbleLens ``Original'' and ``Clean'' sets have an identical number of pages. 

\begin{table}[htbp]
\begin{center}
   \caption{CER on  ScribbleLens full pages. ``Original'' pages vs ``Clean'' pages with an uncertainty of +/- 0.6\%}
   \label{tab:cleanvsorg}

    \begin{tabular}{ | l | c | c| }
    \hline
						& \multicolumn{2}{c|}{ \textbf{Corpus CER [\%]} }	\\ \cline{2-3}
		\textbf{Model}	  & \textbf{Clean}  & \textbf{Original}		\\ \hline
		Clean  		& 9.71	& 11.24  \\
		Original		& 10.06	& 10.62 \\ \hline 
	\end{tabular}
	\vspace{-0.5cm}
\end{center}
\end{table}

In a nutshell, the result in Table~\ref{tab:cleanvsorg} shows that accuracy on the ``Clean'' corpus is marginally but not significantly more accurate compared to the ``Original'' pages. The mis-matched scenarios show an expected, small degradation in accuracy.

Finally, we quantify the effect on accuracy due to the 2D page layout with explicit line wraps. For a sequence of $N$ target symbols on a page with $M$ lines of text, you could argue that the page image is easier to recognize in a long, flat, input image of size $[64 \times W*M]$ or $[64 \times W_1]$, a long one dimensional (1D) flat line of text with no line breaks, compared to the normal page layout of $[64*L \times W_2]$ in a two dimensional way (2D) way which includes line breaks.

The normal 2D page layout can cause hyphenation at end of lines, with words split across two lines, e.g., like ``seg-'' ``mentation''. In this experiment, we compare the 1D and 2D layout and start from the same transcripts. However, in the 1D layout, we do merge the hyphenated words such as ``seg-'' ``mentation'' to ``segmentation''.

There is evidence from~\cite{BlucheLM16,Wigington18} that ``whole paragraph'' recognition with a 2D layout of a set of lines is possible. Our preliminary experiments on synthetic sequences of MNIST digits, laid out in flat, 1D sequences versus the paragraph and page-like 2D layout, showed that these images can be transcribed with the correct target sequence. Line breaks on a page do not seem to cause confusion in the sequence of symbols. While the current experiment is with European languages written left-to-right, we expect that Arabic manuscripts written right-to-left, or traditional Chinese top-to-bottom, are handled with similar grace.

As a reminder, the 1D flattened pages are part of the ScribbleLens~\cite{Openslr19} corpus and constructed by adding line after line to the right. The 2D pages are the original manuscript scans straight from the archives. For RODRIGO, we constructed the 1D flat lines from the individual lines as well.

\begin{table}[htbp]
\begin{center}
   \caption{Accuracy of flat (1D) page input $[64 \times W_1]$ versus normal resized (2D) pages $[64*L \times W_2]$.}
   \label{tab:1d2d}

    \begin{tabular}{| l | c | c|}
    \hline
			 			& \multicolumn{2}{c|}{\textbf{CER [\%]} } \\\cline{2-3}
		\textbf{Corpus} & \textbf{1D}  & \textbf{2D}			 \\
    \hline
		RODRIGO  		&  3.16 +/- 0.11 	& 2.30 +/- 0.09 \\ 
		ScribbleLens  	& 10.35 +/- 0.59 	& 10.62 +/- 0.60 \\ 
    \hline
	\end{tabular}

\vspace{-0.5cm}
\end{center}
\end{table}

For both corpora, there is no catastrophic break-down in accuracy for either 1D page lines or the original 2D pages which is great news. We speculate that training on the whole pages teaches the model about ``noise'' in the margins, ``noise'' from page numbers, and words hyphenated at line end.

\section{Conclusion}
\label{sec:conclusion}

We investigated whole page recognition of historical manuscripts. No explicit line or word segmentation is involved in inference, nor a text localization mechanism. Based on this page-in, text-out approach, we conclude that the whole page, segmentation free recognition from manuscript images is feasible and competitive in terms of CER.

Given enough training data, the page-based accuracy in unconstrained CER is very similar to recognition on line-segmented data.
In edge cases, when words wrap around line ends, the accuracy of a whole page recognizer might even be better as it avoids unrecoverable segmentation errors.
We compared page recognition with and without "wrap around" line ends and see small effects only.

Historical handwritten manuscripts are often high on image noise due to their age.
We showed that manuscript noise such as margin annotations have only a small effect on whole-page recognition accuracy, and certainly no catastrophic degradation. However, having enough transcribed pages to train from is key in supervised training which is why we built ScribbleLens as another research tool for un-, weakly-, and distant- supervised training~\cite{chorowski2019unsupervised,Lancucki2020robust} on historical manuscripts for further research. We firmly believe that representation learning and un-supervised recognition models will have more accuracy and robustness in future, as the amount of digitized but un-transcribed historical pages in the archives is several orders of magnitudes higher compared to transcribed ones.

In terms of software complexity and models, the presented approach of recognition-only without text localization, segmentation or detection is simpler than many ``text-in-the-wild'' OCR systems. Hence, future research could investigate whether it is possible to simplify ``text-in-the-wild'' OCR systems further, and whether these whole-page recognition models can augment or surpass more traditional OCR approaches, especially for older books with more artifacts and less standard fonts.

With respect to other improvements for the model in this paper, the approach of dilated CNNs~\cite{Yu16} was successful in other settings and modalities and they might help on the whole page as well to provide more context. However, our initial results showed no improvement. 

{
\bibliographystyle{IEEEbib}
\baselineskip 10pt
\bibliography{segmentationfree}

\begin{thebibliography}{10}

\bibitem{Nakagawa04}
C.L. Liu, S.~Jaeger, and M.~Nakagawa,
\newblock ``Online recognition of {C}hinese characters: {T}he
  state-of-the-art,''
\newblock {\em IEEE Trans. Pattern Analysis and Machine Intelligence}, vol. 26,
  no. 2, pp. 198–213, 2004.

\bibitem{Steinherz99}
T.~Steinherz, E.~Rivlin, and N.~Intrator,
\newblock ``Off-line {C}ursive {S}cript {W}ord {R}ecognition - {A} {S}urvey,''
\newblock {\em International Journal on Document Analysis and Recognition
  (IJDAR)}, vol. 2, pp. 90--110, 1999.

\bibitem{Vincarelli02}
A.~Vincarelli,
\newblock ``A survey of offline cursive word recognition,''
\newblock {\em Pattern Recognition}, vol. 35, no. 7, pp. 1433--1446, 2002.

\bibitem{Bozinovic89}
R.M. Bozinovic and S.N. Srihari,
\newblock ``Off-line cursive script word recognition,''
\newblock {\em IEEE Trans. Pattern Analysis and Machine Intelligence}, vol. 11,
  no. 1, pp. 68--83, 1989.

\bibitem{Mehri19}
M.~{Mehri}, P.~{Héroux}, R.~{Mullot}, J.~{Moreux}, B.~{Coüasnon}, and
  B.~{Barrett},
\newblock ``{ICDAR2019} competition on historical book analysis - {HBA2019},''
\newblock in {\em 2019 International Conference on Document Analysis and
  Recognition (ICDAR)}, 2019, pp. 1488--1493.

\bibitem{Dolfing98}
J.G.A. Dolfing,
\newblock {\em {H}andwriting {R}ecognition and {V}erification : {A} hidden
  {M}arkov approach},
\newblock Ph.D. thesis, Eindhoven University of Technology (TUE), 1998,
\newblock
  \url{https://research.tue.nl/en/publications/handwriting-recognition-and-verification-a-hidden-markov-approach}.

\bibitem{Senior94}
A.W. Senior,
\newblock ``An off-line cursive handwriting recognition system,''
\newblock {\em IEEE Trans. Pattern Analysis and Machine Intelligence}, vol. 20,
  pp. 309--321, 1998.

\bibitem{Manke94}
S.~Manke and U.~Bodenhausen,
\newblock ``A connectionist approach for on-line cursive handwriting
  recognition,''
\newblock in {\em International Conference on Acoustics, Speech, and Signal
  processing (ICASSP)}, 1994, vol.~2, pp. 633--636.

\bibitem{Starner94}
T.~Starner, R.~Schwartz, J.~Makhoul, and G.~Chou,
\newblock ``On-line cursive handwriting recognition using speech recognition
  methods,''
\newblock in {\em International Conference on Acoustics, Speech, and Signal
  processing (ICASSP)}, 1994, pp. 125--128.

\bibitem{BlucheLM16}
T.~Bluche, J.~Louradour, and R.O. Messina,
\newblock ``Scan, attend and read: End-to-end handwritten paragraph recognition
  with {MDLSTM} attention,''
\newblock {\em CoRR}, vol. abs/1604.03286, 2016.

\bibitem{Ney84}
H.~Ney,
\newblock ``The use of a one-stage dynamic programming algorithm for connected
  word recognition,''
\newblock {\em IEEE Trans. Acoustics, Speech and Signal processing}, vol. 32,
  pp. 263--271, 1984.

\bibitem{Wang18}
D.~Wang, X.~Wang, and S.~Lv,
\newblock ``An {O}verview of {E}nd-to-{E}nd {A}utomatic {S}peech
  {R}ecognition,''
\newblock {\em Symmetry}, vol. 8, pp. 1018, 2019.

\bibitem{Bunke99}
U.V. Marti and H.~Bunke,
\newblock ``A full {E}nglish sentence database for off-line handwriting
  recognition,''
\newblock in {\em International Conference on Document Analysis and Recognition
  (ICDAR))}, 1999, pp. 705 -- 708.

\bibitem{Rodrigo10}
N.~Serrano, F.~Castro, and A.~Juan,
\newblock ``The \uppercase{RODRIGO} database,''
\newblock in {\em Proc. of the 7th International Conference on Language
  Resources and Evaluation (LREC)}, 2010, p. 2709–2712,
\newblock Written 1545.

\bibitem{Dolfing20}
J.G.A Dolfing, J.R. Bellegarda, J.~Chorowski, R.~Marxer, and A.~Laurent,
\newblock ``The ``{S}cribblelens'' {D}utch historical handwriting corpus,''
\newblock in {\em Int. Conference on Frontiers of Handwriting Recognition
  (ICFHR)}, 2020.

\bibitem{Vinciarelli02}
A.~Vinciarelli,
\newblock ``A survey of offline cursive word recognition,''
\newblock {\em Pattern recognition}, vol. 35, no. 7, pp. 1433--1446, 2002.

\bibitem{Voigtlaender16}
P.~Voigtlaender, P.~Doetsch, and H.~Ney,
\newblock ``Handwriting recognition with large multidimensional long short-term
  memory recurrent neural networks,''
\newblock in {\em International Conference on Frontiers in Handwriting
  Recognition (ICFHR)}, 2016, pp. 228--234.

\bibitem{Nephi18}
O.~Nina, R.~Ault, and R~Pack,
\newblock ``Nephi: An open source {P}ytorch library for handwriting
  recognition,''
\newblock in {\em Family History Technology Workshop, Brigham Young
  University}, 2018.

\bibitem{Graves09}
A.~Graves, M.~Liwicki, S.~Fernández, R.~Bertolami, H.~Bunke, and
  J.~Schmidhuber,
\newblock ``A novel connectionist system for unconstrained handwriting
  recognition,''
\newblock {\em IEEE transactions on pattern analysis and machine intelligence},
  vol. 31, pp. 855--68, 06 2009.

\bibitem{Sanchez17}
J.A. Sanchez, V.~Romero, A.H. Toselli, M.~Villegas, and E.~Vidal,
\newblock ``\uppercase{Icdar}2017 competition on handwritten text recognition
  on the \uppercase{READ} dataset,''
\newblock in {\em 14th IAPR International Conference on Document Analysis and
  Recognition (ICDAR)}, 2017, vol.~1, p. 1383–1388.

\bibitem{Ogier08}
J.M. Ogier,
\newblock ``Ancient {D}ocument {A}nalysis : A set of new research problems,''
\newblock in {\em Colloque International Francophone sur l’Ecrit et le
  Document}, 2008, pp. 73--78.

\bibitem{Openslr19}
J.G.A. Dolfing, J.R. Bellegarda, J.~Chorowski, R.~Marxer, and A.~Laurent,
\newblock ``{T}he ``{S}cribblelens'' {D}utch historical handwriting corpus,''
  \url{https://openslr.org/84}, 2019.

\bibitem{Lancucki2020robust}
A.~Łańcucki, J.~Chorowski, G.~Sanchez, R.~Marxer, N.~Chen, J.G.A. Dolfing,
  S.~Khurana, T.~Alumäe, and A.~Laurent,
\newblock ``Robust training of vector quantized bottleneck models,'' 2020.

\bibitem{Shi16}
B.~{Shi}, X.~{Bai}, and C.~{Yao},
\newblock ``An end-to-end trainable neural network for image-based sequence
  recognition and its application to scene text recognition,''
\newblock {\em IEEE Transactions on Pattern Analysis and Machine Intelligence},
  vol. 39, no. 11, pp. 2298--2304, 2017.

\bibitem{Bartz18}
C.~Bartz, H.~Yang, and C.~Meinel,
\newblock ``See: Towards semi-supervised end-to-end scene text recognition,''
\newblock in {\em AAAI Conference on Artificial Intelligence}, 2018,
\newblock \url{http://arxiv.org/abs/1712.05404}.

\bibitem{Moysset17}
B.~{Moysset}, C.~{Kermorvant}, and C.~{Wolf},
\newblock ``Full-page text recognition: Learning where to start and when to
  stop,''
\newblock in {\em 14th IAPR International Conference on Document Analysis and
  Recognition (ICDAR)}, 2017, vol.~01, pp. 871--876.

\bibitem{Wigington18}
C.~Wigington, C.~Tensmeyer, B.~Davis, W.~Barrett, B.~Price, and S.~Cohen,
\newblock ``Start, follow, read: End-to-end full-page handwriting
  recognition,''
\newblock in {\em The European Conference on Computer Vision (ECCV)}, 2018, pp.
  376--383.

\bibitem{Jaderberg16}
M.~Jaderberg, K.~Simonyan, A.~Vedaldi, and A.~Zisserman,
\newblock ``Reading text in the wild with convolutional neural networks,''
\newblock {\em Int. Journal Computer Vision}, vol. 116, pp. 1--20, 2016.

\bibitem{IAM-HistDB}
A.~Fischer, E.~Indermühle, H.~Bunke, Viehauser, and M.~Stolz,
\newblock ``Ground truth creation for handwriting recognition in historical
  documents,''
\newblock in {\em Proc. 9th Int. Workshop on Document Analysis System (DAS)},
  2010, pp. 3--10.

\bibitem{Germana10}
D.~P\'{e}rez, L.~Taraz\'{o}n, N.~Serrano, F.~Castro, O.~Ramos, and A.~Juan,
\newblock ``The \uppercase{GERMANA} database,''
\newblock in {\em Proc. of the 10th International Conference on Document
  Analysis and Recognition (ICDAR)}, 2010, p. 301–305.

\bibitem{READ}
``Recognition and enrichment of archival documents (\uppercase{READ}),''
  \url{https://read.transkribus.eu}, 2016-2019.

\bibitem{Washington41}
G.~Washington,
\newblock ``George {W}ashington papers at the {L}ibrary of {C}ongress, series
  2, letterbook 1, pages 270-279 and 300-309,''
\newblock 1741-1799.

\bibitem{Wal15}
M.~van~der Wal, G.~Rutten, J.~Nobels, and T.~Simons,
\newblock ``{T}he letters as loot/ {B}rieven als buit corpus ({L}eiden
  {U}niversity),'' \url{http://brievenalsbuit.inl.nl}, 2015,
\newblock {T}ranscriptions, metadata and images of manuscripts from the 17th
  and 18th centuries.

\bibitem{EMMO}
H.~Wolfe,
\newblock ``{E}arly {M}odern {M}anuscripts {O}nline (\uppercase{EMMO}),
  {F}olger {S}hakespeare {L}ibrary,'' \url{https://emmo.folger.edu}, 2017,
\newblock {T}ranscriptions, metadata and images of manuscripts from the 16th
  and 17th centuries.

\bibitem{CASIAAHCDB19}
Y.~{Xu}, F.~{Yin}, D.~{Wang}, X.~{Zhang}, Z.~{Zhang}, and C.~{Liu},
\newblock ``{CASIA}-{AHCDB}: {A} {L}arge-{S}cale {C}hinese {A}ncient
  {H}andwritten {C}haracters {D}atabase,''
\newblock in {\em International Conference on Document Analysis and Recognition
  (ICDAR)}, 2019, p. 793–798.

\bibitem{Granell18}
E.~Granell, E.~Chammas, L.~Likforman-Sulem, C.-D. Martin\'{e}z-Hinarejos,
  C.~Mokbel, and B.-I. C\^{\i}rstea,
\newblock ``Transcription of \uppercase{S}panish historical handwritten
  documents with deep neural networks,''
\newblock {\em Journal of Imaging}, pp. 1--22, 4(1), 15, 2018.

\bibitem{Bunke02}
U.V. Marti and H.~Bunke,
\newblock ``The {IAM}-database: An {E}nglish {S}entence {D}atabase for
  {O}ff-line {H}andwriting {R}ecognition,''
\newblock {\em Int. Journal on Document Analysis and Recognition}, vol. 5, pp.
  39--46, 2002.

\bibitem{IAMSplits99}
``{IAM} data sets splits,''
  \url{http://www.fki.inf.unibe.ch/DBs/iamDB/tasks/largeWriterIndependentTextLineRecognitionTask.zip},
  1999.

\bibitem{Scheidl18}
H.~Scheidl,
\newblock ``Deslanting algorithm,''
  \url{https://github.com/githubharald/DeslantImg}, 2018.

\bibitem{Such16}
F.~P. {Such}, D.~{Peri}, F.~{Brockler}, H.~{Paul}, and R.~{Ptucha},
\newblock ``Fully convolutional networks for handwriting recognition,''
\newblock in {\em 16th International Conference on Frontiers in Handwriting
  Recognition (ICFHR)}, 2018, pp. 86--91.

\bibitem{Pham13}
V.~{Pham}, T.~{Bluche}, C.~{Kermorvant}, and J.~{Louradour},
\newblock ``Dropout improves recurrent neural networks for handwriting
  recognition,''
\newblock in {\em 14th International Conference on Frontiers in Handwriting
  Recognition}, 2014, pp. 285--290.

\bibitem{Kozielski13}
M.~Kozielski, P.~Doetsch, and H.~Ney,
\newblock ``Improvements in {RWTH}’s system for off-line handwriting
  recognition,''
\newblock in {\em 12th International Document Analysis and Recognition
  (ICDAR)}, 2013, pp. 935--939.

\bibitem{Dreuw11}
P.~Dreuw, P.~Doetsch, C.~Plahl, and H.~Ney,
\newblock ``Hierarchical hybrid {MLP/HMM} or rather {MLP} features for a
  discriminatively trained gaussian {HMM}: {A} comparison for offline
  handwriting recognition,''
\newblock in {\em IEEE Int. Conference on Image Processing (ICIP)}, 09 2011,
  pp. 3541--3544.

\bibitem{Bleda11}
S.~España-Boquera, M.J. Castro-Bleda, J.~Gorbe-Moya, and F.~Zamora-Martínez,
\newblock ``Improving offline handwritten text recognition with hybrid hmm/ann
  models,''
\newblock {\em IEEE Trans. on Pattern Analysis and Machine Intelligence}, vol.
  33, pp. 767--79, 04 2011.

\bibitem{Poznanski16}
A.~{Poznanski} and L.~{Wolf},
\newblock ``{CNN-N-G}ram for handwriting word recognition,''
\newblock in {\em IEEE Conference on Computer Vision and Pattern Recognition
  (CVPR)}, 2016, pp. 2305--2314.

\bibitem{chorowski2019unsupervised}
J.~Chorowski, N.~Chen, R.~Marxer, H.~Dolfing, A.~{\L}a{\'n}cucki, G.~Sanchez,
  T.~Alum{\"a}e, and A.~Laurent,
\newblock ``Unsupervised neural segmentation and clustering for unit discovery
  in sequential data,''
\newblock in {\em NeurIPS 2019 workshop-Perception as generative
  reasoning-Structure, Causality, Probability}, 2019.

\bibitem{Specaug2019}
D.S. Park, W.~Chan, Y.~Zhang, C.C. Chiu, B.~Zoph, E.D. Cubuk, and Q.V. Le,
\newblock ``Spec{A}ugment: {A} {S}imple {D}ata {A}ugmentation method for
  automatic speech recognition,''
\newblock {\em Interspeech 2019}, Sep 2019.

\bibitem{Chorowski19}
J.~Chorowski, N.~Chen, R.~Marxer, J.G.A. Dolfing, A.~{\L}a{\'n}cucki,
  G.~Sanchez, T.~Alum{\"a}e, and A.~Laurent,
\newblock ``Distsup: A framework for unsupervised and distant-supervised
  representation learning with variational autoencoders (vq-vae, som-vae,
  etc),'' \url{https://github.com/distsup/DistSup}, 2019.

\bibitem{Seytre19}
J.~Seytre, J.~Wu, and A.~Achille,
\newblock ``{T}ext{T}ubes for {D}etecting {C}urved {T}ext in the {W}ild,''
  arXiv:1912.08990.

\bibitem{Vaswani17}
A.~Vaswani, N.~Shazeer, N.~Parmar, J.~Uszkoreit, L.~Jones, A.N. Gomez, \L.
  Kaiser, and I.~Polosukhin,
\newblock ``Attention is all you need,''
\newblock in {\em Advances in Neural Information Processing Systems 30}, pp.
  5998--6008. 2017.

\bibitem{TransformerEncoder20}
PyTorch,
\newblock ``Transformerencoder,''
  \url{https://pytorch.org/docs/master/generated/torch.nn.TransformerEncoder.html#transformerencoder},
  2020.

\bibitem{STN}
M.~Jaderberg, K.~Simonyan, A.~Zisserman, and K.~Kavukcuoglu,
\newblock ``Spatial transformer networks,''
\newblock in {\em Advances in Neural Information Processing Systems 28}, 2015,
  pp. 2017--2025.

\bibitem{Curriculum09}
Y.~Bengio, J.~Louradour, R.~Collobert, and J.S. Weston,
\newblock ``Curriculum {L}earning,''
\newblock in {\em ICML}, 2009, pp. 41--48.

\bibitem{Guyon96}
I.~Guyon, R.~Schwartz, J.~Makhoul, and V.~Vapnik,
\newblock ``What size test set gives good error rate estimates?,''
\newblock in {\em IEEE Trans. PAMI}, 1996, pp. 52--64.

\bibitem{Yu16}
F.~Yu and V.~Koltun,
\newblock ``Multi-scale context aggregation by dilated convolutions,''
\newblock in {\em Int. Conference on Learning Representations (ICLR)}, 2016.

\end{thebibliography}
}
\end{document}